\pdfoutput=1

\documentclass[11pt]{article}

\usepackage[]{acl}

\usepackage{times}
\usepackage{latexsym}

\usepackage[T1]{fontenc}

\usepackage[utf8]{inputenc}
\usepackage{times}
\usepackage{soul}
\usepackage{url}
\usepackage{graphicx}
\usepackage{amsmath}
\usepackage{amsthm}
\usepackage{multirow}
\usepackage{booktabs}
\usepackage{algorithm}
\usepackage{algorithmic}
\usepackage{microtype}
\title{PSP: Pre-trained Soft Prompts for Few-Shot Abstractive Summarization}

\author{Xiaochen Liu$^{\dag}$, \textbf{Yang Gao$^{\dag}$}\thanks{~Corresponding author.}, \textbf{Yu Bai$^{\dag}$},\\ \textbf{Jiawei Li$^{\dag}$},   \textbf{Yinan Hu$^{\dag}$}, \textbf{Heyan Huang$^{\dag}$} \and \textbf{Boxing Chen} \\ $^{\dag}$School of Computer Science and Technology, \\ Beijing Institute of Technology \\ \texttt{\{xcliu,gyang,yubai,jwli,ynhu,hhy63\}@bit.edu.cn}\\ \texttt{chenboxing@gmail.com}
}
\begin{document}
\maketitle
\begin{abstract}
Few-shot abstractive summarization has become a challenging task in natural language generation. To support it, we developed a novel soft prompts architecture coupled with a prompt pre-training plus prompt fine-tuning paradigm, which is effective and tunes only extremely light parameters. 
To meet the structure of the generation models, the soft prompts comprise continuous input embeddings across an encoder and a decoder. Importantly, a new inner-prompt placed in the text is introduced to capture document-level information. 
The aim is to devote attention to understanding the document that better prompts the model to generate document-related content. 
In the training process, the prompt pre-training with self-supervised pseudo-data firstly teaches the model basic summarizing capability. Then, with few-shot examples, only the designed lightweight soft prompts are fine-tuned. 
Experimental results on the CNN/DailyMail and XSum datasets show that our method, with only 0.1\% of the parameters, outperforms full-model tuning where all model parameters are tuned. It also surpasses Prompt Tuning by a large margin and delivers competitive results against Prefix-Tuning with 3\% of the parameters.
\end{abstract}

\section{Introduction}

Given the high labor-costs of obtaining quality abstractive summaries, few-shot abstractive summarization is very demanding and highly challenging. 
A widely accepted paradigm for almost all NLP tasks is to fine-tune the entire set of parameters for a large pre-trained language model to suit the target task~\citep{liu2019text,liu2020multilingual}.

However, the fine-tuning with few-shot examples usually leads to disappointing results, especially with generation tasks like abstractive summarization~\citep{fabbri2020improving,yu2021adaptsum}. The likely outcome is an overfit model. 
Further, for every specific task, a large number of pre-trained parameters need to be updated and stored, which is not efficient to use.

Pre-trained language models are few-shot learners, i.e., GPT-3 \citep{brown2020language} that surprisingly perform generation tasks from a few examples without any further gradient updates. Although it lacks a rigorously theoretical proof, prompt learning inherits the few-shot
property \citep{li-liang-2021-prefix,schick2020few,jin2021good,liu2021gpt}. Commonly, this type of learning is considered to retrieve relevant knowledge from frozen language models, only tuning continuous prompts to quickly adapt to new tasks with very few examples.

More recently, Prompt Tuning~\citep{lester-etal-2021-power} has received much attention. With large frozen language models (say, $>$10 billion parameters), Prompt Tuning simply adds a tunable soft prompt to the input of the encoder, achieving results that are comparable to full-model tuning. Yet, our empirical results, in Section \ref{section:pilot}, demonstrate that Prompt Tuning for abstractive summarization yields simply abysmal performance. 
Prefix-Tuning~\citep{li-liang-2021-prefix} extends the use of prompt learning in the natural language generation area. With this technique, continuous prompts are applied to every layer of the pre-trained model and even shows increase in few-shot generation tasks  over fine-tuning. 
Yet the training process is not stable and updates are required that add to the memory and training costs.\footnote{See more related work in Section \ref{sec:related work_appendix}.}

\begin{figure}[t]
    \centering
    \includegraphics[width=0.48\textwidth]{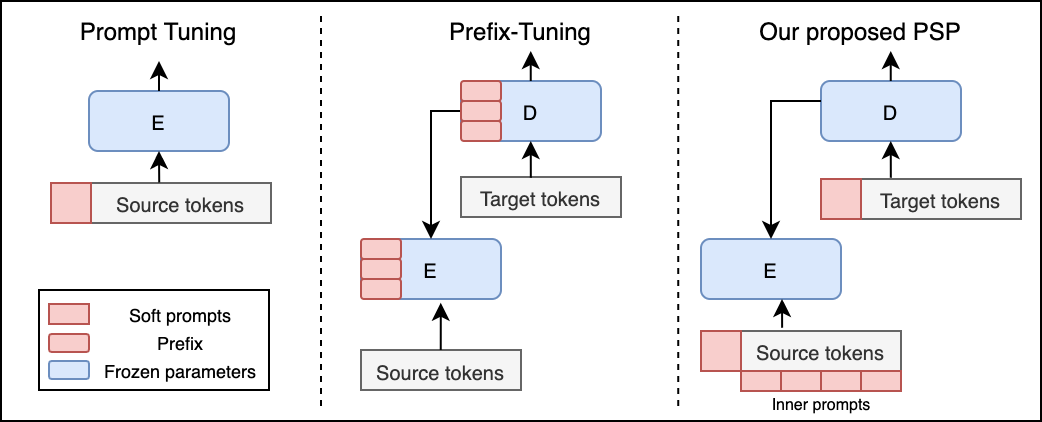}
    \caption{The comparison between PSP and previous methods. ``E'' and ``D'' represents the encoder and the decoder, respectively.}
    \label{fig:intro}
\end{figure}

Given the shortcomings of these two methods, we have developed a soft prompts tuning method that is specifically designed for summarization. The structure is given in Figure \ref{fig:intro}. 
The method is capable of performing few-shot language generation task (i.e., abstractive summarization) with an efficient amount of training parameters. 
Prompt tokens are added before the decoder input tokens to guide the generation process toward the target summary. Moreover, we have designed three inner prompts – interval, sequential, and fixed-length – one of which is placed among the source input tokens. 
The aim is to capture the structure in the source document and aid in understanding its semantics, 
so as to better prompt the model to generate document-related content. Each kind of inner prompts
focuses on different semantic units (e.g., phrases, sentences, and etc.),  differentiating important units from non-informative ones. 
To bolster the summarization ability of the model and assist the prompts to understand the documents, prompt pre-training is performed before the tuning process, and leveraged by self-supervised pseudo data. As a last step, all the prompts are fine-tuned with few-shot training examples.
Experiments conducted on two commonly used datasets - CNNDM~\citep{see2017get} and XSum~\citep{xsum-emnlp} - demonstrate that our  method outperforms full-model tuning under few-shot settings only with 0.1\% of the parameters. It also surpasses naive Prompt Tuning by a large margin. Our model also yields a performance competitive to Prefix-Tuning with 3\% of the trainable parameters. 
A detailed analysis shows that the designed prompt-pre-training phase and the inner prompts are effective for few-shot text summarization. Thus, the major contributions of this work include :
1) A novel soft prompt architecture for few-shot abstractive summarization. With the well-designed prompts in embedding layer, our model fulfills the task effectively and efficiently; 
2) It is necessary to perform prompt pre-training strategy which benefits soft prompts model for few-shot summarization and shows excellent zero-shot capabilities; 
3) Experiments that investigate the effect of different prompts by probing the attention weights. The results show our model is able to: extract knowledge from the encoder language model; understand the discourse in the document; and guide the decoder language model to generate fluent summaries.

\section{Pilot Experiments}\label{section:pilot}

In a pilot study, we experimented with using Prompt Tuning under 300-shots settings to find reasonable clues as to how to design summary-prompts for the task. Our findings follow.

Consider an encoder-decoder language model $p_{\theta}(y|x)$ based on the Transformer architecture~\citep{Vaswani2017AttentionIA} (e.g., BART~\citep{lewis2020bart}) and parameterized by $\theta$. 
To conduct a few-shot summarization task, we have some few-shot training pairs of a document  $X = \{x_1, x_2, \dots, x_{|X|}\}$ and a corresponding summary $Y= \{y_1, y_2, \dots, y_{|Y|}\}$. Specifically, we divided $X$ into different subsets with \textbf{sentences}\footnote{Note that, throughout this work, a 
``sentence'' can be an arbitrary span of contiguous text (e.g., fixed length of 10 tokens), or an actual linguistic sentence.} as our unit,  $X = \{x^1_1, \dots x^i_j,\dots, x^n_m \}$, where $x^i_j$ denotes the $j_{\rm th}$ token in the $i_{\rm th}$ sentence.

First, original Prompt Tuning is applied by concatenating a series of prompt tokens ${P}_{en}$, parameterized by $\theta_{p_{en}}$,  to the encoder input $X_{en} = \{e^1_1, \dots, e^i_j, \dots e^n_m\}$, where $e$ represents the embedding of each token (the leftmost structure in Figure~\ref{fig:intro}). The gradients are backpropagated through the prompts and the weights $\theta$ of language model are frozen~\cite{lester-etal-2021-power}. In this way, the model maximizes the likelihood of the output $Y$:
\begin{equation}
    p_{\theta;\theta_{\tt p_{en}}}(Y|[P_{en};X_{en}])
\end{equation}%
The result of original Prompt Tuning is shown on the first line in Table \ref{tab:pt variants}, where we see it severely underperforms versus full-model tuning.  
In further experiments, we added a series of prompts $P_{de}$ to the decoder inputs $X_{de}$ following the generation $p_{\theta;\theta_{p_{de}}}(Y|X_{en},P_{de})$. Here, we found the results to be even worse than the last.

\paragraph{Necessary Prompts for Generation} For generation-based tasks, prompts in both the encoder and decoder are equivalently useful. Therefore, our model  employs a combination of the two series of prompts mentioned above, and generates $Y$ conditioning on $X_{en}$, $P_{en}$ and $P_{de}$: 
\begin{equation}
    p_{\theta;\theta_{p_{en}};\theta_{p_{de}}}(Y|[P_{en};X_{en}],P_{de})
\end{equation}

\begin{table}
\centering
\resizebox{.9\linewidth}{!}{
\begin{tabular}{lccc}
\toprule
{Model}  & {ROUGE-1} & {ROUGE-2} & {ROUGE-L}  \\
\midrule
Prompt in encoder & 32.87  & 11.92  & 21.73    \\
Prompt in decoder  & 26.77  & 11.73  & 16.71    \\
Prompt in en.\&de. & 36.37 & 14.41 &\textbf{24.46}  \\
Full-Model Tuning  & \textbf{37.01}  & \textbf{14.49} &  23.91  \\
\bottomrule
\end{tabular}
}
\caption{Results of BART-base on CNN/DailyMail Datasets. Best results are bold.
}
\label{tab:pt variants}
\end{table}
The result on the third line in Table~\ref{tab:pt variants}  again verify our hypothesis. Prompts across the encoder and decoder even achieve comparable results with full-model tuning under few-shot settings. This verifies two things for us. First, prepending simple prompts to only the input embedding layer is effective and efficient for few-shot abstractive summarization. Second, prompts across the encoder and decoder are both necessary for generation tasks.  
\begin{figure}[t]
    \centering
    \includegraphics[width=0.45\textwidth]{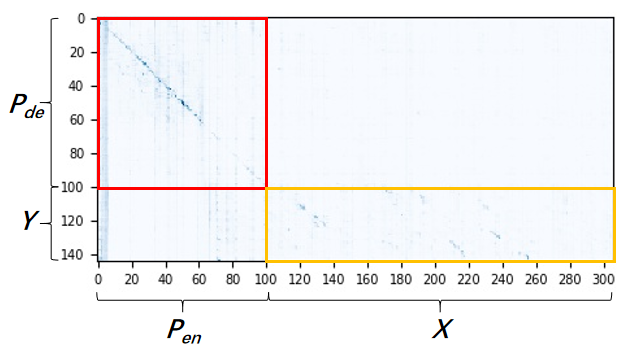}
    \caption{Visualization of the encoder-decoder attention weights. The x-axis are the encoder input, including prompts across the encoder $P_{en}$ and the source document $X$. The y-axis are the decoder input, including prompts across the decoder $P_{de}$ and the target summary $Y$. The area in the red box represents the attentions of $P_{de}$ assigning to $P_{en}$. The area in the yellow box represents the attentions of $Y$ assigning to $X$. Darker color shows the more highly related associations between tokens.}
    \label{fig:nopretrain_noinner_attn}
\end{figure}

\paragraph{Lack of Attention on the Document} 
We further explored the encoder-decoder attention to investigate the effect of the prompts and freezing the language model. From Figure \ref{fig:nopretrain_noinner_attn}, we find the generating output is mainly focused on the soft prompts to come with little attention given to the document itself. 
This outcome is detrimental to summarization that requires to understand the semantics and inner discourse structure of documents~\cite{wang2019concept}. Without the associations of target summaries and source documents, it is impossible to obtain high-quality summaries using current prompt architectures. 

From Figure \ref{fig:nopretrain_noinner_attn}, we can observe that prompts in the encoder and the ones in decoder are consistently and directly associated with each other. We speculate that the mechanism is that encoder prompts retrieve relevant knowledge from the frozen encoder language model as a document representation, and decoder prompts copy the encoder's behaviour, guiding the decoder language model to generate text.  




\begin{figure}[t]
    \centering
    \includegraphics[width=0.50\textwidth]{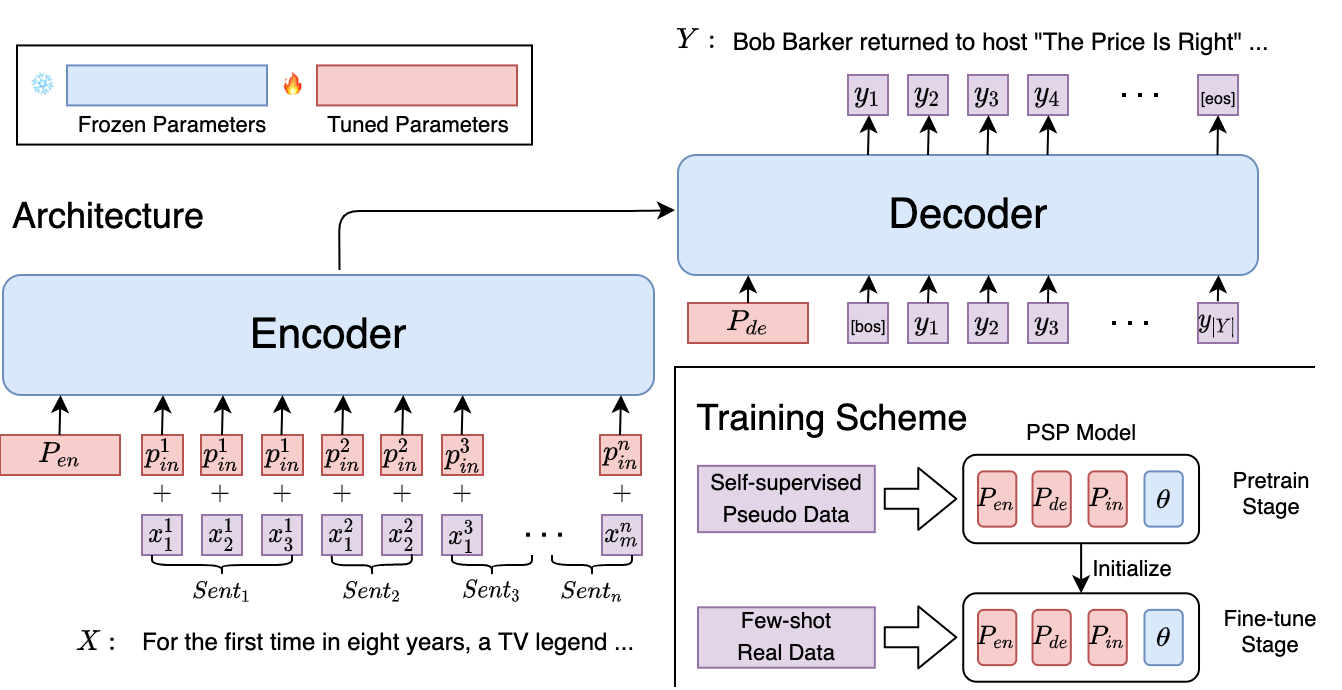}
    \caption{Architecture and training scheme of PSP. Squares in blue and red indicates frozen and tuned parameters, respectively. }
    \label{fig:arch}
\end{figure}

\section{Method}\label{section:method}
In light of our findings about the current architectures, we developed a new architecture of pre-trained soft prompts, for few-shot abstractive summarization called PSP. The framework includes continuous prompts across the encoder and decoder inputs, as well as inner-prompts to capture the dependencies between documents and target summaries.
To better understand a given document, we add a prompt pre-training process before few-shot tuning. It also brings a good initialization for the prompting. The overall architecture and training scheme are illustrated in Figure~\ref{fig:arch}.   




\subsection{Encoder-Decoder Basic Prompts}
As mentioned in Section~\ref{section:pilot}, 
in the training phase of current architectures, $P_{en}$ is responsible for extracting knowledge from the encoder's frozen language model as a document representation. Meanwhile, $P_{de}$ mostly copies the behavior of $P_{en}$ and guides the frozen decoder's language model to generate fluent text as a summary. 

To strengthen the model’s ability to understand a document, the dependencies and attentions given to the source document need to be embodied in the prompt architecture.



\subsection{Inner-Prompts for Document Understanding}


To achieve our goal, we propose the notion of adding inner-prompts within the source document, denoted as $P_{in}= \{p_{in}^{1}, p_{in}^{2}, \dots, p_{in}^{n}\}$ with the parameters $\theta_{P_{in}}$ to be updated. Each $p_{in}^i$ corresponds to a single sentence. These inner-prompts are added to the corresponding token embedding, which gives rise to a new $X'_{in}$:
\begin{equation}
\small
X'_{in} = \{e^1_1 + p^1_{in}, e^1_2 + p^1_{in}, \dots, e^i_j + p^{i}_{in}, \dots, e^n_m + p^n_{in} \}
\end{equation}

We believe that by prompting different semantic units (e.g., sentences, phrases, etc.), more attention can be given to understanding the document’s discourse. Furthermore, the inner-prompts help the model to quickly interpret the document by strengthening the associations between outputs and documents. 
What follows are three different strategies for incorporating the three different inner-prompts. Note that there is more discussion on this point in Section \ref{sec:analysis on inner}.

\begin{figure}[t]
    \centering
    \includegraphics[width=0.50\textwidth]{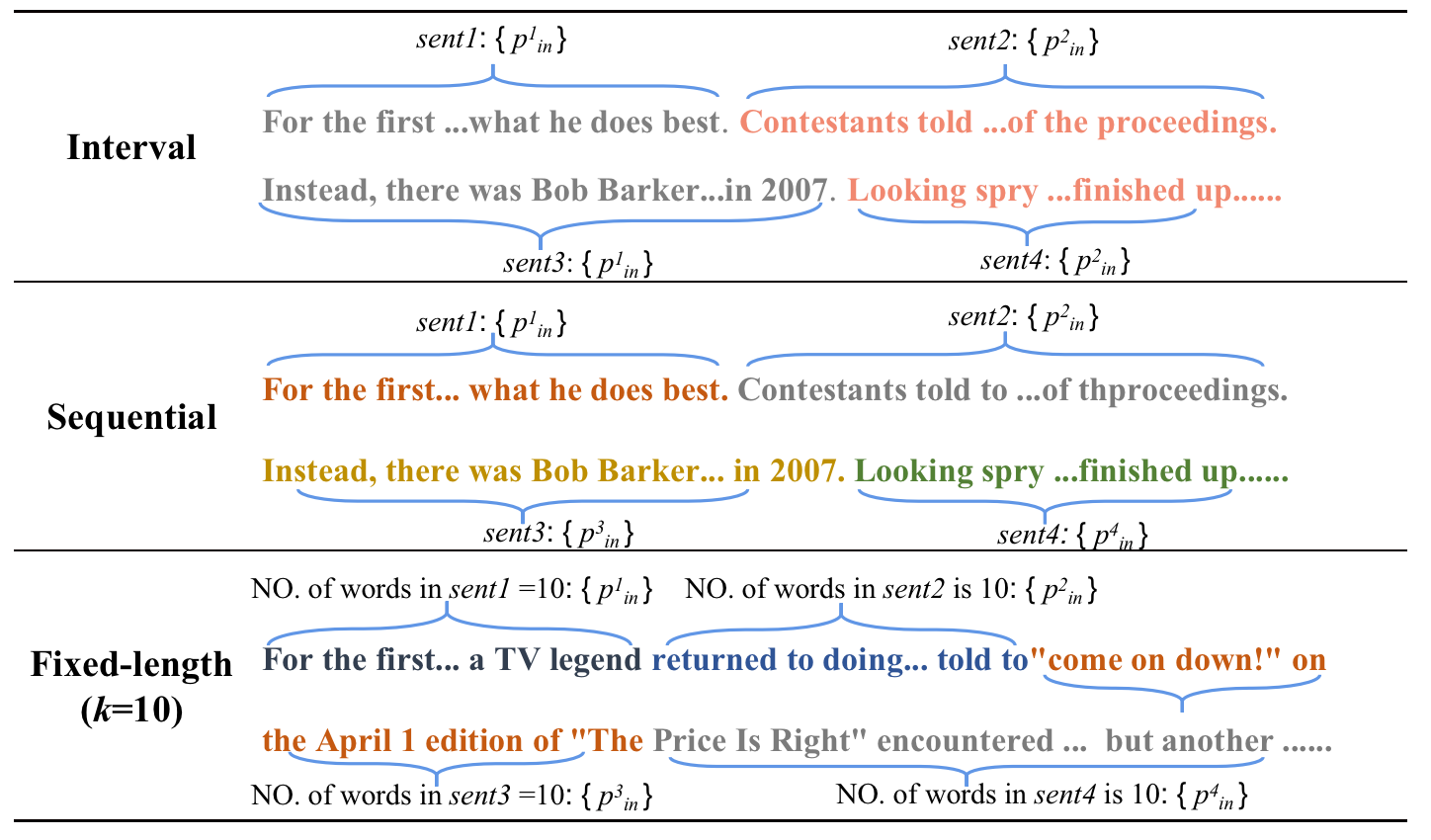}
    \caption{Different inner prompts for one example source document. Different colors indicate different inner prompt embeddings. ``NO. of words'' means the length of the text span.}
    \label{fig:inner}
\end{figure}
 
\paragraph{Interval} Following \citet{liu2019text}, the interval inner-prompts comprises two inner-prompt tokens are assigned to each sentence $sent_i$, depending on whether $i$ is odd.
Specifically,
\begin{equation}
P_{in}= \{p_{in}^{1}, p_{in}^{2}, p_{in}^{1}, \dots, p_{in}^{(n-1) {\rm mod} 2+1}\}
\end{equation}
In this way, the model can identify important sentences to encode the document at sentence level.
\paragraph{Sequential} To highlight the complex discourse structure of documents, sentence positions need to be considered. Therefore, different tokens are set in sentences by their sequences, formulated as: 
\begin{equation}
    P_{in}= \{p_{in}^{1}, p_{in}^{2}, \dots, p_{in}^{n}\}
\end{equation}
\paragraph{Fixed-length} To discover more fine-grained semantic units, a text span with a fixed length $k$ is manipulated into a new ``sentence'' and a corresponding sequential token is assigned to it.
Further, prompts are assigned to the newly divided sentences [$sent_{1}$, $sent_{2}$, ..., $sent_{n}$], as $\{p_{in}^{1}, p_{in}^{2}, \dots, p_{in}^{n}\}$.
Figure~\ref{fig:inner} illustrates some examples where the above strategies have been used.

\subsection{Self-supervised Prompt Pre-training}
To improve ability of the prompts to understand the documents and to help the model to adapt to the summarization tasks, soft prompts are further pre-trained on the corpus using summarization-oriented self-supervised objectives. Doing this also means that the prompts are well initialized for few-shot tuning. 

We tested two strategies for constructing the self-supervised data. Each strategy was designed to suit a particular type of writing bias in the document. These are ``lead” and ``gap sentences generation”.

\paragraph{Lead} Lead bias is common in news articles, which usually follow an inverted pyramid structure where the first few sentences contain the most salient information~\citep{see2017get,yang2020ted}. 
With this type of bias, we initially select the first three sentences as our target summary, and treated the rest of the document as the source text. 
With this type of prompt pre-training process, the model was able to infer the salient information based on the remaining text.

\paragraph{GSG} Gap sentences generation applies to all documents that do not follow the lead bias structure (e.g., XSum~\citep{xsum-emnlp}).
The strategy used here follows \citet{zhang2020pegasus}
, where we used ROUGE1-F1~\citep{lin2004rouge} between each sentence $x_{i}$ and the
rest of the document as a proxy for the principal score, $s_{i}=rouge(x_{i},D \setminus \{x_{i}\}),\forall{i}$. 
The top-$m$ most important sentences were selected according to $s_i$, and removed from the document. Then these $m$ sentences are concatenated in the same order as the original text in the form of a pseudo summary. The remainder of the text is treated as a pseudo document. 

With the constructed data, our designed prompts can be pre-trained and further tuned with few-shot examples.

\subsection{Training Objective}
The model is trained with maximum likelihood estimation (MLE). Given a ground-truth summary $Y = [y_{1}, y_{2}, ..., y_{|Y|}]$ for an input passage $X$, the objective is to minimize the negative log-likelihood of the target word sequence:
\begin{equation}
\begin{aligned}
   \mathcal{L} = -\sum_{t=1}^{|Y|}\log p_{\theta^{*}}(y_{t}|[P_{en};X'_{in}],[P_{de};y_{1},...y_{t-1}])\\
   \theta^{*} = \{\theta;\theta_{p_{en}};\theta_{p_{de}};\theta_{p_{in}}\}
   \end{aligned}
\end{equation}
Note that only these prepended-prompts parameters ($\theta_{p_{en}}$, $\theta_{p_{de}}$) and the inner-prompts parameters ($\theta_{p_{in}}$) are optimized, the language model parameters ($\theta$) are all frozen.


\section{Experiments}

\paragraph{Datasets} We experimented with the CNN/DailyMail (CNNDM) dataset~\citep{hermann2015teaching} and the XSum dataset~\citep{xsum-emnlp}. We chose these datasets because they differ in abstraction level  and text length, which helps to show the generalization ability of our results. 

We constructed the self-supervised pre-training data for CNNDM with Lead, and for XSum with GSG. We show details in Section \ref{sec:pseudo data_appendix} in the appendix. 
Given that the lead bias structure exists only in some domain-specific datasets, we also conducted experiments to demonstrate the universality of the GSG to construct pseudo-data. The results are shown in Section~\ref{sec:universality of gsg_appendix} in the appendix.
Our few-shot training set $D_{train}$ contained 300 document-summary pairs randomly sampled from the original training data. To tune the hyper-parameters and select the best checkpoint, we composed a validation set $D_{dev}$ from the original validation data. Here, we were careful to ensure that $\lvert D_{train} \rvert = \lvert D_{dev} \rvert$ so that it fit into a true few-shot learning setting, following \citet{perez2021true}. Since few-shot learning may have high variance, we sampled the examples with 5 different random seeds. We used the original test set to report our results, including the mean value and the standard deviation. Table \ref{tab:Datasets statistics} shows the statistics of the pre-processed corpus.

\begin{table}[t]
\centering
\small
\resizebox{.8\linewidth}{!}{
\begin{tabular}{lrrrrrr}
\toprule
\multirow{2}*{Datasets}  & \multicolumn{3}{c}{CNNDM}  & \multicolumn{3}{c}{XSum}  \\
\cmidrule(r{4pt}){2-4} \cmidrule{5-7}
~ & train & dev & test & train & dev & test \\
\midrule
Avg.Passage  & 697.45 &676.64 &717.92  & 396.53&387.62&380.55      \\
Avg.Sum     & 55.91 &51.97 &58.62  & 22.90&23.29&22.11       \\
Labled data  & 300&300&11,490  & 300&300&11,333      \\
\bottomrule
\end{tabular}}
\caption{Datasets statistics. “Avg.Passage'' means the average length of passages and “Avg.Sum” means the average length of summaries.}
\label{tab:Datasets statistics}
\end{table}

\paragraph{Setup} The base version of BART was used in our work. Following \citet{lester-etal-2021-power}, we used 100 prompt tokens for both the encoder inputs and the decoder inputs. These prompts were randomly initialized from the set of vocabularies. 
The sequential and fixed-length inner-prompts require a maximum number. 
Hence, we counted the number of sentences in each document and divided the results into two groups – the 85\% with the least sentences (Group A) and the 15\% with the most sentences (Group B)\footnote{We made our division at 85\% to ensure all embeddings of inner-prompt tokens could be fully trained, because sentences after the $n$-th only exist in 15\% of the data.}. 
We then set the number of prompts to the most number of sentences in Group A plus one, i.e., $n + 1$. For CNNDM, that number was 61 and, for XSum, it was 33.  In this way, one inner-prompt token was assigned to each sentence up to $n$. For the excessively long documents in Group B, the text after $n$ sentences was assigned an $n + 1$-th token.
Further, we drew from a normal distribution $\mathcal{N}(0,0.05)$ to initialize the inner-prompt embeddings\footnote{More information about implementation details are shown in Section~\ref{sec:implementation details_appendix} in the appendix.}. Taking CNNDM as an example, all the tunable parameters that need to be stored amount to only $2\times 10^5$.
This is compared to the ($1.4\times10^8$) parameters of full-model tuning. That equates to around 0.1\% of the parameters for each dataset that need to be tuned and stored.
\paragraph{Evaluation Metrics} We adopted  ROUGE~\cite{lin2004rouge} to measure the quality of the summaries produced in our experiments. The F1 scores for ROUGE-1, ROUGE-2, and ROUGE-L between the ground-truth and the generated summaries are each reported.
\paragraph{Baseline Models} We compared PSP to: {\bf Prompt Tuning}~\citep{lester-etal-2021-power}, which only concatenates soft prompts into the encoder input; {\bf Prefix Tuning}~\citep{li-liang-2021-prefix}, which adds a prefix to all the encoder layers, cross-attention layers, and the decoder layers; and {\bf Full-Model Tuning}, which does not have any prompts and fine-tunes all the parameters of the pre-trained language model.

\begin{table*}[t]
\begin{center}
\resizebox{.9\linewidth}{!}{
\begin{tabular}{llrrrrrrrr}
\toprule
\multicolumn{2}{c}{}                        & \multicolumn{4}{c}{CNNDM} & \multicolumn{4}{c}{XSum} \\
\multicolumn{2}{l}{Model}                   & ROUGE-1             & ROUGE-2             & ROUGE-L             & PPL     & ROUGE-1            & ROUGE-2         & ROUGE-L   & PPL \\ \midrule
\multicolumn{2}{l}{Prompt Tuning}            & $30.58_{2.07}$   & {$11.93_{0.46}$}  & $21.73_{1.86}$          & $141.56$   & $29.63_{1.21}$  & $8.84_{0.55}$    & $22.00_{1.23}$ & $101.96$\\
\multicolumn{2}{l}{Prefix-Tuning}            & $37.12_{0.15}$   & ${\bf 16.59_{0.09}}$  & ${\bf 26.28_{0.06}}$  & $52.59$  & $32.18_{0.16}$  & $11.13_{0.08}$  & $25.50_{0.14}$  & $39.58$ \\
\multicolumn{2}{l}{Full-Model Tuning}       & $38.03_{0.56}$   & $16.01_{0.79}$      & $25.21_{0.70}$             & $65.73$ & $32.85_{0.25}$  & $10.52_{0.24}$  & $25.15_{0.29}$ & $51.63$ \\ \midrule
\multicolumn{2}{l}{PSP$_{\tt{Interval}}$}                & $37.82_{0.29}$   & {$15.40_{0.31}$}  & $25.10_{0.36}$  &${\bf 45.54}$  & $\underline{\bf 32.86_{0.21}}$  & $\underline{\bf 11.27_{0.08}}$  & $\underline{\bf 25.64_{0.11}}$ & $44.25$\\
\multicolumn{2}{l}{PSP$_{\tt{Sequential}}$}               & $37.82_{0.39}$   & {$15.58_{0.32}$}  & $25.16_{0.32}$ & $48.10$ & $32.57_{0.11}$  & $\underline{10.97_{0.07}}$  & $\underline{25.39_{0.05}}$  &${\bf 35.70}$\\
\multicolumn{2}{l}{PSP$_{\tt{Fixed-k}}$}                  & $\underline{\bf 38.31_{0.15}}$   & $15.94_{0.21}$  & $\underline{25.41_{0.25}}$  & $58.50$ & $32.81_{0.10}$  & $\underline{11.15_{0.10}}$  & $\underline{25.48_{0.13}}$  & $52.10$ \\ \bottomrule
\end{tabular}   
}
\end{center}
\caption{Results on CNNDM and XSum Datasets. The experiments are conducted with 300 training samples and 300 validation samples on each dataset. We report the mean value and the standard deviation over 5 sampled datasets. $k$ = 10 is chosen for {PSP$_{\tt{Fixed-k}}$}. ``PPL'' represents the perplexity of generated summaries. A low perplexity indicates the summaries are fluent. Best results are bold and underline means our models outperform Full-model tuning. }
\label{tab:automatic evaluation}
\end{table*}

\subsection{Experimental Results of Our Method}
Table~\ref{tab:automatic evaluation} presents the results of all PSP variants and baselines across CNNDM and XSum datasets.
With the exception of the ROUGE-2 and ROUGE-L scores for the Prefix-Tuning on the CNNDM dataset, our proposed PSP, outperforms the others. However, PSP delivered a competitive result with only 3\% of the parameters, which is an acceptable place to start.
To our surprise, we observe that 50\% of PSP's results surpass the full-model tuning, especially on XSum, as underlined in the table.
Besides, results on the PPL metric show that PSP can generate more fluent summaries than other models. These results indicate that fine-tuning large language models is not necessarily a good or efficient idea with few-shot generation. It also shows that soft prompts with frozen language models are effective for few-shot abstractive summarization. 
Moreover, it statistically verifies that PSP with its three inner-prompt strategies is effective.

\paragraph{Efficiency v.s. effectiveness.} 
We  gave an overall comparison to baseline models on effectiveness and memory-efficiency, evaluated by ROUGE and the number of parameters, respectively. The results are shown in Table~\ref{tab:efficient}. Prompt Tuning has the least number of parameters, while its capacity is limited to this and lacks control over the decoder side, hence it can not perform natural language generation tasks well. We can see that substantial gains are made when going from vanilla Prompt Tuning to PSP. However, even if Prefix-Tuning is nearly thirty times more parameters than ours, there is either a marginal improvement or even performance decrease on some metrics. Besides, Prefix-Tuning relies on reparameterization tricks to stabilize the training, i.e., adds a MLP with large number of parameters to the training stage. Our method provides the best effectiveness-efficiency trade off, and outperforms full-model tuning with only 0.1\% parameters, and presents competitive results against Prefix-Tuning with 3\% parameters. 

\begin{table}[t]
\centering
\small
\resizebox{.9\linewidth}{!}{
\begin{tabular}{lcccc}
\toprule
\multirow{2}{*}{Model} & \multirow{2}{*}{\# Train} & \multirow{2}{*}{\# Store} & \multicolumn{2}{c}{ROUGE-1} \\ \cmidrule(r{4pt}){4-4} \cmidrule{5-5}
                        &                           &                           & CNNDM        & XSUM         \\ \midrule
PSP                     & $2.0\times10^5$          & $2.0\times10^5$            & {\bf 38.32}        & {\bf 32.86}        \\
Prefix-Tuning           & $2.4\times10^7$           & $5.5\times10^6$       & 37.12        & 32.18        \\
Prompt Tuning           & $7.7\times10^4$           & $7.7\times10^4$       & 30.58        & 29.63        \\
Full-Model Tuning       & $1.4\times10^8$       & $1.4\times10^8$       & 38.03           & 32.85        \\ \bottomrule
\end{tabular}}
\caption{Comparison with baseline models on effectiveness and efficiency. ``\# Train'' means the number of tuned parameters during training. `` \# Store'' means the number of stored parameters. Best results are bold.}
\label{tab:efficient}
\end{table}

\paragraph{Human Evaluation}
We conducted a human evaluation study. To this end, we randomly selected 20 instances from the test set of each dataset. Ten graduate students with high levels of fluency in English were asked to assess the generated summaries and golden summaries from independent perspectives~\cite{wang2021exploring}: {\it Informativeness} (how much useful information does the summary provide?), {\it Relevance} (how well does the summary reflect the input document?), and {\it Fluency} (how grammatically correct are the summary sentences and how easy are they to read?). Scoring followed the Best-Worst Scaling method~\citep{kiritchenko2017best}. Participants were asked to select the best and worst summaries from each perspective. The scores were computed as the percentage of times a summary was chosen as the best minus the times it was selected as the worst. The scores ranged from -1
(worst) to 1 (best). Results are shown in Table~\ref{tab:human evaluation}. Qualitatively, we show several examples generated by different models and the reference in Table~\ref{tab:cases cnndm} and Table~\ref{tab:cases xsum} in the appendix. Compared with all baselines, the summaries generated by PSP are always more fluent and relevant to the source document, consistent with the results of human evaluation. Further more, we found summaries generated by PSP and Prefix-Tuning are always similar in sentence patterns and expressions. However, Prefix-Tuning tends to generate texts shorter than PSP, which often leads to lack of information. 

\begin{table}[t]
\centering
\small
\resizebox{.9\linewidth}{!}{
\begin{tabular}{lrrrrrr}
\toprule
\multirow{2}*{Methods}  & \multicolumn{3}{c}{CNNDM}  & \multicolumn{3}{c}{XSum}  \\
\cmidrule(r{4pt}){2-4} \cmidrule{5-7}
~ & IF & RL & FL & IF & RL & FL \\
\midrule
PSP  & {\bf 0.500}&{\bf 0.708}&{\bf 0.667}  & {\bf 0.217}&{\bf 0.275}&{\bf 0.492}      \\
Prompt Tuning     & -0.317&-0.758&-0.975  & -0.336&-0.400&-0.867       \\
Prefix-Tuning  & -0.233&0.067&0.158  & 0.017&-0.008&0.292      \\
Full-Model Tuning  &0.067 &-0.025&0.075  & 0.117&0.092&0.075      \\ \bottomrule
\end{tabular}}
\caption{Human evaluation results. Best results are bold.}
\label{tab:human evaluation}
\end{table}

\paragraph{Selection of fixed length $k$.}
As shown in Table~\ref{tab:automatic evaluation}, {PSP$_{\tt{Fixed-k}}$} performs consistently well on both datasets. So we further explored the influence of different length $k$, i.e., $k=5, 10, 15, 30$, for inner-prompt tokens of the PSP$_{\tt Fixed-k}$\footnote{The average number of tokens per sentence in both datasets was about 18, so we did not consider fixed lengths of 20, for its similarity to the PSP$_{\tt Sequential}$.}. Table~\ref{tab:fixed length} presents the results of the variants on XSum. 
We observe the segmented spans with 10 tokens achieve the best performance. Interestingly, it can be induced that, to understand a document, it is possible to reorganize the sentence into several semantic units, where the number of the tokens is 10 on average.
We also report results of different $k$ on our validation set in Table~\ref{tab:fixed length}. The ranking is consistent with the test set. From a practical perspective, when applying PSP to a new dataset, we can choose the best $k$ based on the validation set.
\begin{table}[t]
\centering
\resizebox{0.9\linewidth}{!}{
\begin{tabular}{lrrrrrr}
\toprule
\multirow{2}*{$k$} & \multicolumn{3}{c}{$D_{dev}$} & \multicolumn{3}{c}{$D_{test}$} \\
\cmidrule(r{4pt}){2-4} \cmidrule{5-7}
                   & R-1    & R-2    & R-L   & R-1    & R-2    & R-L    \\ \midrule
5                  & 34.27  & 11.90   & 26.41 & 31.90   & 10.28  & 24.20   \\
10                 & {\bf 35.31}  & {\bf 12.88}  & {\bf 26.85} & {\bf 32.89}  & {\bf 11.13}  & {\bf 25.51}  \\
15                 & 34.98  & 11.68  & 26.45 & 32.11  & 10.46  & 24.72  \\
30                 & 34.48  & 12.57  & 26.55 & 32.20   & 11.03  & 25.30   \\ \bottomrule
\end{tabular}
}
\caption{Results of different fixed length $k$ on validation set $D_{dev}$ and test set $D_{test}$ of XSum. ``R-1'' is short for ``ROUGE-1'', the same for ``R-2'' and ``R-L''.}
\label{tab:fixed length}
\end{table}

\subsection{Analyses on Soft Prompts}
\label{sec:analysis on inner}
\begin{figure}[ht]
    \centering
    \includegraphics[width=0.5\textwidth]{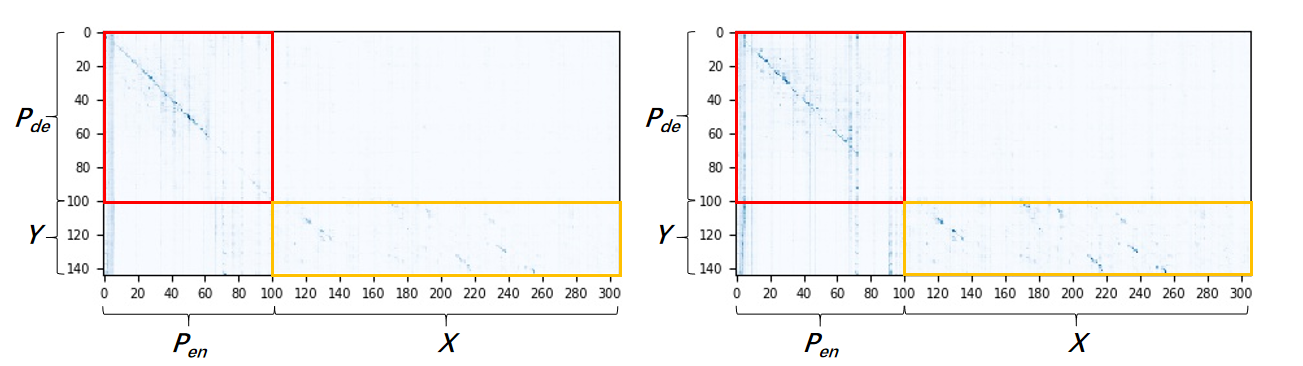}
    \caption{Visualization of the encoder-decoder attention weights of the model with only prompts across the encoder and the decoder (left) and PSP (right). Detailed descriptions refer to Figure \ref{fig:nopretrain_noinner_attn}.}
    \label{fig:pretrain+inner_attn}
\end{figure}

\paragraph{Whether our model attends to understand documents?} 
According to Figure \ref{fig:nopretrain_noinner_attn}, we further present the encoder-decoder attention distribution of the PSP. The comparison visualization is shown in Figure~\ref{fig:pretrain+inner_attn}. 
We find the following enhancement of our model by introducing the inner prompts. First, the PSP model strengthens the associations between the encoder prompts and the decoder prompts compared to the original model.
Second, the soft prompt $P_{en}$ has more opportunities to be related to the output $Y$, indicating the semantic relations between them. 
Third, the output $Y$ assigns more attention to the source document $X$. 
This suggests that the hidden structure of the document is  emphasized, increasing the capability of understanding its semantics. 
As such, these prompts can properly elect  salient information from the document and prompt the model to generate the output.  

\paragraph{Do inner prompts assist the model to understand the content of documents or simply increase the model's capacity?}
Instead of using inner-prompts, we prepended additional tunable tokens (i.e. 150 tokens) in front of the encoder and the decoder inputs. Comparison results are shown in Table~\ref{tab:analyses on inner}. Despite the larger capacity, soft prompts with 150 tunable tokens before the input performed the worst,  denoted as \textit{soft prompts (en.\&de., 150)}.  This suggests the inner-prompts 
with a few parameters do help to understand the document by prompting the structures, rather than simply add more trainable parameters to increase the model's capacity. 

\begin{table}[t]
\centering
\small
\resizebox{\linewidth}{!}{
\begin{tabular}{lrrrrrr}
\toprule
\multirow{2}*{Model} & \multicolumn{3}{c}{CNNDM} & \multicolumn{3}{c}{XSum} \\
\cmidrule(r{4pt}){2-4} \cmidrule{5-7}
                              & R-1    & R-2    & R-L   & R-1    & R-2    & R-L    \\ \midrule
Soft prompts (en.\&de., 100)            & 36.89   & 14.96   & 24.63  & 29.36   & 9.90   & 22.92  \\
Soft prompts (en.\&de., 150)             & 35.71    & 14.86   & 23.97 &28.94   &9.52   &22.24  \\ 
Soft prompts (en.\&de.\&ip., 100)     & {\bf 37.87}   & {\bf 15.83}   & {\bf 25.37}     & {\bf 31.95} &{\bf 10.52} &{\bf 24.80} \\ \bottomrule
\end{tabular}}
\caption{Results of different architectures of soft prompts on CNNDM and XSum, where ``en.'' ``de.'' ``ip.'' are short for encoder, decoder and inner prompts, respectively. Numbers in parentheses represent the number of prompt tokens we prepended before the encoder and decoder input.}
\label{tab:analyses on inner}
\end{table}

\begin{table}[t]
\small
\begin{center}
\resizebox{\linewidth}{!}{
\begin{tabular}{lrrr}
\toprule
  Model & ROUGE-1 & ROUGE-2 & ROUGE-L \\ \midrule
Soft prompts (en.\&de., shared) & 36.06   & 14.30   & 24.24   \\
Soft prompts (en.\&de., separate)  & 36.37   & 14.41   & 24.46        \\ \bottomrule
\end{tabular}
}
\end{center}
\caption{Results of basic soft prompts on the CNNDM.
}
\label{tab:share}
\end{table}
\paragraph{Further insight on soft prompts across the encoder and the decoder.}
To verify our hypothesis that the decoder prompts largely copy the behaviour of the encoder prompts, we shared similar embeddings of the soft prompts before the encoder and the decoder. In Table~\ref{tab:share}, we observe the Soft prompts (en.\&de., shared) and (en.\&de., separate) almost perform identical results. Although the parameters are only half of the original model, the performance consistently remains competitive. 
This shows that the shared prompts can extract important information from the document and further guide the language model to generate consistently good summaries more efficiently.

\begin{figure}[t]
    \centering
    \includegraphics[width=0.5\textwidth]{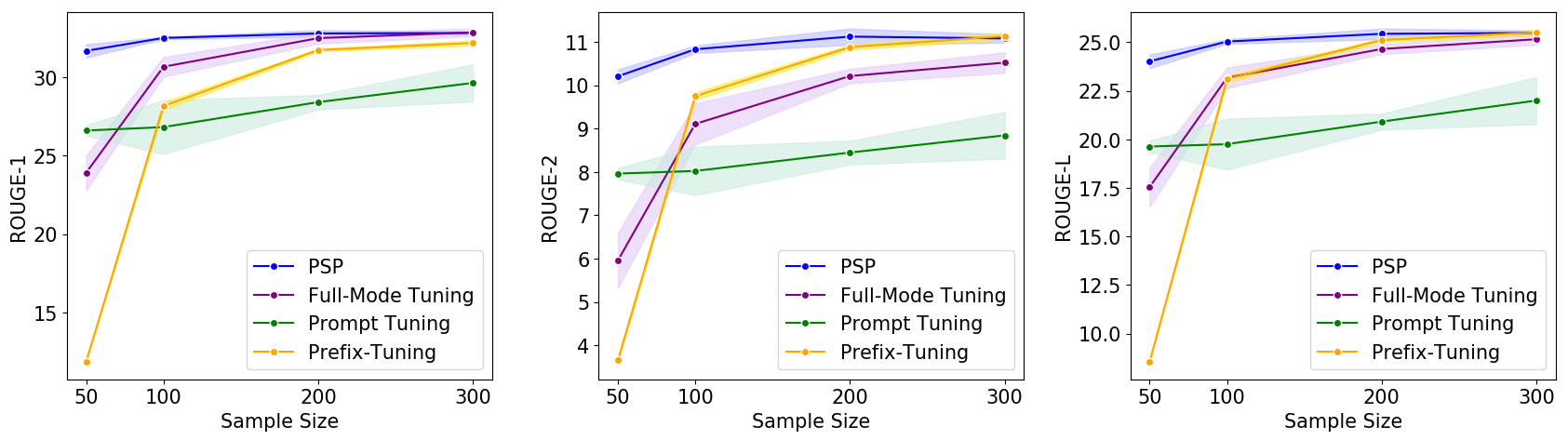}
    \caption{$k$-shot summarization results on XSum.}
    \label{fig:xsum kshot}
\end{figure}

\begin{table}[t]
\small
\begin{center}
\resizebox{.95\linewidth}{!}{
\begin{tabular}{lrrr}
\toprule
Model  & ROUGE-1 & ROUGE-2 & ROUGE-L \\ \midrule
Full-Model Tuning  & 11.69   & 2.67   & 7.74   \\
Prefix-Tuning  & 11.76   & 2.63   & 7.93       \\ 
Prompt Tuning  & 9.40   & 1.86   & 6.19       \\ 
PSP$_{\tt{Interval}}$  & {\bf 17.16}   & {\bf 3.36}   & {\bf 12.65}       \\ \bottomrule
\end{tabular}
}
\end{center}
\caption{Zero-shot results on XSum.
}
\label{tab:zero-shot}
\end{table}

\subsection{Analysis on Few-shot and Zero-shot Summarization}
To examine the performance of different methods under few-shots, we further randomly sampled number of \{50, 100, 200\} as the settings. Figure~\ref{fig:xsum kshot} reports a more detailed overview of all models' performance across
a range of different few-shots. The ROUGE scores of our model generally outperform other baselines and remain steady across different scenarios. Especially, the PSP with only 50 examples receives the most significant improvements, while the Prefix-Tuning doesn't even work (tuning based on BART$_{\tt base}$) possibly due to its instability of the model.  
Moreover, we report the results of zero-shot on XSum in Table~\ref{tab:zero-shot}. Benefiting from the knowledge gained in the pre-training phase, our model shows a significant advantage of  zero-shot adaptation in generating quality summaries.  

\subsection{The Performance of Pre-training on Prefix-Tuning}
A crucial strategy for PSP is the pre-training of soft prompts. To give a fairly comparison, we performed prefix pre-training for Prefix-Tuning in the same way with the PSP. The results are shown in Table~\ref{tab:pretrain prefix}. We can find that the Prefix model obtains improvements on the XSum dataset after adopting the pre-training strategy,  but underperforms the original one on the CNNDM dataset. It indicates that Prefix-Tuning shows limited potential compared to our model. We induce that the pre-training for Prefix-Tuning raises over-fitting risk due to its sensitivity to different data or parameter settings. 
%

\begin{table}[t]
\centering
\resizebox{\linewidth}{!}{
\begin{tabular}{lrrrrrr}
\toprule
\multirow{2}*{Method} & \multicolumn{3}{c}{CNNDM}   & \multicolumn{3}{c}{XSum}    \\
\cmidrule(r{4pt}){2-4} \cmidrule{5-7}
~                      & ROUGE-1 & ROUGE-2 & ROUGE-L & ROUGE-1 & ROUGE-2 & ROUGE-L \\ \midrule
Prefix-Tuning         & $37.12_{0.15}$   & ${16.59_{0.09}}$  & ${26.28_{0.06}}$  & $32.18_{0.16}$  & $11.13_{0.08}$  & $25.50_{0.14}$  \\ 
Prefix-Tuning w/ Pre. & $37.35_{0.58}$  & $16.08_{0.37}$  & $25.95_{0.50}$ & $33.39_{0.10}$  & $11.61_{0.06}$  & $26.07_{0.09}$ \\ \bottomrule
\end{tabular}}
\caption{Test set results of Prefix-Tuning. ``w/ Pre.'' means that we pre-trained the prefix with pseudo data.}
\label{tab:pretrain prefix}
\end{table}

\subsection{Ablation Study}
We conducted experiments to examine the effectiveness of the major components of our model, and Table \ref{tab:ablation_methods} shows the ablation results across the two datasets. 
We observed both the prompt pre-training operation and the inner-prompts component contribute to the main model. Notably, with the removal of each component, the model becomes considerably unstable, indicated by the variance shown in the ablation results.
Comparably, prompt pre-training in our model accounts for more importance on the XSum dataset whose summaries have a higher abstract level (we assume it's more ``difficult'') than the CNNDM.
In sum, these two components support the performance and stability of our model in terms of summarization adaption (by prompt pre-training) and structural documents understanding (by inner-prompts).  

\begin{table}[ht]
\small
\centering
\resizebox{\linewidth}{!}{
\begin{tabular}{lrrrrrr}
\toprule
\multirow{2}*{Method} & \multicolumn{3}{c}{CNNDM}   & \multicolumn{3}{c}{XSum}    \\
\cmidrule(r{4pt}){2-4} \cmidrule{5-7}
~ & ROUGE-1 & ROUGE-2 & ROUGE-L & ROUGE-1 & ROUGE-2 & ROUGE-L \\ \midrule
{PSP$_{\tt{Fixed-k}}$}          & $38.31_{0.15}$   & $15.94_{0.21}$  & $25.41_{0.25}$  & $32.81_{0.10}$  & $11.15_{0.10}$  & $25.48_{0.13}$ \\
\quad w/o PP                       & {$37.30_{0.56}$}  & $15.45_{0.39}$  & $24.93_{0.38}$  & $32.17_{0.16}$  & $10.69_{0.13}$ & $25.02_{0.21}$ \\
\quad w/o IP                       & $37.76_{0.28}$   & {$15.22_{0.31}$}  & $24.80_{0.40}$  & $32.59_{0.17}$  & $11.14_{0.17}$  & $25.46_{0.24}$ \\
\quad w/o PP \& IP        & $36.88_{0.42}$   & {$14.96_{0.45}$}  & $24.63_{0.40}$  & $29.35_{1.5}$  & $9.87_{0.43}$  & $22.89_{1.19}$ \\ \bottomrule
\end{tabular}}
\caption{Ablation study of PSP on two datasets. ``w/o'' means without. ``PP'' and ``IP'' are short for Prompt Pre-training and Inner-Prompts, respectively. The variance of each result is provided. }
\label{tab:ablation_methods}
\end{table}

\section{Related Work}
\label{sec:related work_appendix}
\paragraph{Few-Shot Abstractive Summarization} 
In practical application scenarios, the lack of manual constructed document-summary pairs or labeled data makes data-driven neural models performs badly~\cite{hu2021gradient, hu2020selfore}. 
~\citet{fabbri2020improving} condense characteristics of the target dataset into Wikipedia data to construct pseudo-summaries. ~\citet{bravzinskas2020few}
introduce plug-in networks to reproduce characteristics of the target dataset with only a small set of labeled examples. ~\citet{bai-etal-2021-cross} conduct cross-lingual summarization in a low-resource setting. ~\citet{yu2021adaptsum} design the second phase of pre-training on
large-scale generative models before fine-tuning. In this paper, we construct pseudo-summary corpus with heuristic rules, providing a better parameter initialization for soft prompts under few-shot settings. More importantly, we design summarization-oriented soft prompts to help the model produce few-shot summaries.


\paragraph{Prompt Learning}
The emergence of GPT-3~\citep{brown2020language} introduces the concept of {\it ``prompting''}. One only needs to assemble a task description and few examples into a prompt, and then prepend it to the task input. With the large-scale frozen parameters, a pre-trained model can generate the output without any task-specific tuning. However, task description is error-prone while there is no unified, explicit, and effective way to build these hard prompts manually~\citep{logan2021cutting}. 
Hence, several works~\citep{gao2020making,jiang2020can,shin2020autoprompt} are proposed to generate prompts automatically, but they all restrict prompts to discrete spaces. These discrete prompts are less expressive and sub-optimal. To overcome the shortcomings of hard prompts,~\citet{li-liang-2021-prefix} propose ``Prefix-Tuning''. This method only tunes prefix activation prepended to all transformer layers, and keeps the LM parameters frozen. To further simplify, Prompt Tuning~\citep{lester-etal-2021-power} only prepends tunable tokens to the encoder input, and keeps all other parameters frozen. 
~\citet{logan2021cutting} and ~\citet{gu2021ppt} propose to use pre-training to boost the low performance of Prompt Tuning for few-shot learning. 
In this work, we fit the structure of Prompt Tuning to text generation models, proposing encoder prompts, decoder prompts, and inner prompts. We successfully apply prompt tuning methods to few-shot abstractive summarization task.
\section{Conclusion}
In this paper, we present a novel pre-trained soft prompts architecture (PSP) specifically designed for few-shot abstractive summarization. We design continuous input embeddings across an encoder and a decoder alongside several kinds of inner-prompts placed in the text, assisting the model better to understand documents and guide accurate generation. Empirical results find the necessity of using prompt pre-training for few-shot/zero-shot abstractive summarization.  
Extensive experiments and analyses show that the proposed PSP provides the best effectiveness-efficiency trade off among all the baseline methods.

\section{Acknowledgments}
The research presented in this publication was sponsored by CCF  Fund For Young Scholars, and Joint Funds of the National Natural Science Foundation of China (Grant No. U21B2009).
\bibliography{anthology,custom}
\bibliographystyle{acl_natbib}

\clearpage
\appendix

\section{Appendix}
\label{sec:appendix}
\subsection{Constructing Pesudo Data for Pre-training}\label{sec:pseudo data_appendix}
We constructed the pseudo data for CNNDM with {\bf Lead}. We also conducted a simple data cleaning procedure to the self-supervised pre-train corpus. 
First, we cleaned away irrelevant information,  such as media names, reporter names or dates from the summaries. 
Second, for those summaries with less than 50 tokens, we iteratively collected the first sentence of the remaining text to the pseudo summary, until the length of summary reaches 70. 
This procedure was set up to prevent the target text from being too short to form a meaningful summary. 
Third, for those samples in which the source document is shorter than its summary, we filtered them out.  

For XSum, we constructed the pseudo data for pre-training following {\bf GSG}. The top-1 most important sentence was selected as the pseudo summary. Then we filtered out those pseudo summaries that are not relevant enough to the pseudo passages. In particular, we leveraged hand-written summaries in the few-shot dataset to determine the filtering threshold of pseudo data. We calculated the ROUGE-1 F1 between each ground-truth summary and its corresponding passage, represented as ${Ri}$. Then we calculated the mean and variance of ${Ri}$: $\epsilon = \frac{1}{n}\sum_{i=1}^nRi$, $\sigma^2 = \frac{1}{n}\sum_{i=1}^n(Ri-\epsilon)^2$, and $\epsilon-\sigma^2$ was used as a lower-bound threshold to filter out low quality pseudo data. For those pseudo samples where ROUGE1-F1 between the pseudo summary and the pseudo passage is lower than the threshold $\epsilon-\sigma^2$, we filtered them out. Finally, we conducted pre-training on our soft prompts with these filtered pseudo-data. Table~\ref{tab:pseduo-data-statistics} shows the statistics for the pre-training data corpus.
\begin{table}[t]
\begin{center}
\resizebox{.9\linewidth}{!}{
\begin{tabular}{cclcl}
\toprule
\multirow{2}{*}{}       & \multicolumn{2}{c}{CNNDM}         & \multicolumn{2}{c}{XSum}          \\
                        & \multicolumn{2}{c}{Pseudo Corpus} & \multicolumn{2}{c}{Pseudo Corpus} \\ \midrule
\# of Original Passages & \multicolumn{2}{c}{287,113}       & \multicolumn{2}{c}{204,017}       \\
\# of Pre-training Data & \multicolumn{2}{c}{284,177}       & \multicolumn{2}{c}{158,499}       \\ \bottomrule
\end{tabular}}
\end{center}
\caption{Pseudo-summarization corpus statistics. ``\# of Original Passages'' means the number of original passages in the training set, ``\# of Pre-training data'' means the number of pseudo data after data cleaning.}
\label{tab:pseduo-data-statistics}
\end{table}

\subsection{Implementation Details}
\label{sec:implementation details_appendix}
 We first split sentences with the Stanford CoreNLP toolkit ~\citep{manning2014stanford}, and the input documents were truncated to 1024 BPE tokens. We adopted BART-base for all the experiments. Our implementation was based on the Hugging Face Transformer models~\citep{wolf2020transformers}. We used a mini-batch size of 8 with a gradient accumulation for 10 iterations. We used Adam optimizer with momentum $\beta_1$ = 0.9, $\beta_2$ = 0.998 and noam decay. In the stage of pre-training, the peak value of learning rate was 1e-3, and we set the warm up ratio to 10\%. During fine-tuning, the peak value of learning rate was 3e-4, and we set the warm up steps to 100 with 400 epochs. In the decoding stage, we used beam search with a beam size of 4. The decoding process will not stop until an end-of sequence (EOS) token was emitted or the length of the generated summary reached to 256 tokens. All models were trained on 4 TITAN RTX GPUs.


\subsection{The Universality of GSG to Construct Pseudo-data}
\label{sec:universality of gsg_appendix}
To demonstrate the universality of using the GSG method to construct pseudo-data for prompt pre-training, we conducted a complimentary experiment to testify its effect on the CNNDM\footnote{We do not conduct ablation experiments on XSum, as there is no `` lead bias'' in this dataset. So it is inappropriate to take the first sentences of the passage as the pseudo summary.}. Specifically, we selected $m=3$ important sentences. Results in Table~\ref{tab:ablation_pseudo} indicate that the PSP model pre-trained by GSG is equally effective with the original PSP$_{\tt Lead}$, showing that the GSG can be universally employed to pre-train soft prompts for abstractive summarization.

\begin{table}[t]
\begin{center}
\centering
\resizebox{.9\linewidth}{!}{
\begin{tabular}{lrrr}
\toprule
  & ROUGE-1 & ROUGE-2 & ROUGE-L \\ \midrule
PSP$_{\tt Lead}$ (w/o inner-prompts)  & 37.66   & 15.07   & 24.52   \\
PSP$_{\tt GSG}$  (w/o inner-prompts)  & 37.04   & 15.04   & 25.20        \\ \bottomrule
\end{tabular}
}
\end{center}
\caption{Results on CNNDM by using the Lead and the GSG to  construct  pseudo-data for prompt pre-training.}
\label{tab:ablation_pseudo}
\end{table}

\begin{table*}
\centering
\resizebox{\linewidth}{!}{
\begin{tabular}{lp{17cm}}
\toprule
Source & Manchester City are monitoring Juventus striker Alvaro Morata. The Spain international was a target for Arsenal a year ago when he left Real Madrid but opted for a move to Italy instead. Real have a buy-back option set at around £15million but are unlikely to take that up. Manchester City are keeping tabs on Juventus striker Alvaro Morata ahead of possible summer bid . Morata closes down Lazio goalkeeper Federico Marchetti during Juve's Serie A win on Saturday . City also remain keen on Juventus midfielder Paul Pogba but face competition from Paris Saint Germain . Morata has been a success at Juventus but the Italians are always susceptible to offers at the right price and are making plans to bring in Edinson Cavani from Paris Saint Germain despite the French club's resistance. PSG also are one of the frontrunners to sign Juve midfielder Paul Pogba. City remain one of the clubs pushing to sign the France international too and Juventus have City's Edin Dzeko, Aleksandar Kolarov and Stevan Jovetic on their list of potential targets for the summer.     \\ \midrule
PSP  & Manchester City are keeping tabs on Juventus striker Alvaro Morata . The Spain international was a target for Arsenal a year ago when he left Real Madrid but opted for a move to Italy instead . Juventus have a buy-back option set at around £15million but are unlikely to take that up . \\
Prefix-Tuning  & Manchester City are keeping tabs on Juventus striker Alvaro Morata ahead of possible summer bid. Morata closes down Lazio goalkeeper Federico Marchetti during Juve's Serie A win on Saturday. \\
Prompt Tuning & LMorata is a target for Real Madrid, who have a buy - back option set at around £ 15million . He has been a success at Real Madrid but opted for a move to Italy instead . The Italian club are unlikely to take that up . . .. . . ." . .." . ." ." ." . ."." . .". . ."". . ." . . " . . ... . . [and] . . (He] had a great season at Real but who more had been about” . . , but who else had been close enough to sign him? . . He had her all after." . .) . . She didn't year when when when two over people A lot more time than two theret. . ." —LastJustInOther \\
Full-Model Tuning & NEW City are monitoring Juventus striker Alvaro Morata . The Spain international was a target for Arsenal a year ago when he left Real Madrid . Morata has a buy-back option set at around £15million but is unlikely to be offered by the French side .\\
Reference & Alvaro Morata had attracted interest from Arsenal before joining Juventus . Spain international made move to Italy in £15million deal from Real Madrid . Manchester City are monitoring the 22-year-old after impressive season .\\ \toprule
Source  & Nairobi, Kenya (CNN)University of Nairobi students were terrified Sunday morning when they heard explosions -- caused by a faulty electrical cable -- and believed it was a terror attack, the school said. Students on the Kikuyu campus stampeded down the halls of the Kimberly dormitory, and some jumped from its fifth floor, the university said. Hundreds were injured and were taken to hospitals. One person died, according to the school. The confusion and panic came less than two weeks after Al-Shabaab slaughtered 147 people at a college in Garissa, Kenya. Kenyan teachers and students have said they fear being targeted by the Somalia-based terrorists. On Sunday, as many as 108 students from the University of Nairobi were admitted to Kenyatta National Hospital. Among them, at least 63 students have been discharged, and at least four are slated for surgery, the school said. Almost all of the 54 students being treated at PCEA Kikuyu Hospital have been released, the university said. Kenya Power authorities and its CEO are at the school and looking into the electrical issue. Normal power supply will resume after repairs, the university said. "As we mourn the unfortunate loss of the departed student, we are also praying for the quick recovery of those who were injured," said Vice Chancellor Peter M.F. Mbithi in a statement. He called on the students, staff and public to remain calm. CNN's Lillian Leposo reported from Nairobi and Ashley Fantz wrote this story in Atlanta.       \\ \midrule
PSP   & University of Nairobi students were terrified Sunday morning when they heard explosions -- caused by a faulty electrical cable -- and believed it was a terror attack . Hundreds were injured and were taken to hospitals . One person died, according to the school . The confusion and panic came less than two weeks after Al- Shabaab slaughtered 147 people at a college in Garissa .\\
Prefix-Tuning & University of Nairobi students were terrified Sunday morning when they heard explosions -- caused by a faulty electrical cable -- and believed it was a terror attack.\\
Prompt Tuning & L Hundreds were injured and taken to hospitals . Hundreds of students were evacuated from the dormitory . The school said they feared being targeted by al-Shabaab . Hundreds were evacuated and treated in hospitals . The university said it is investigating the cause of the explosion . . . The explosion was caused by a faulty electrical cable. . .. . ." . . ." ." ." . ."." . .." . .""People were terrified," said the school's vice chancellor . "People were screaming, but who more had been were about” . "We had no idea what was going on but who else had been about to blow her all after." ... .. ." .."." ..""They were terrified at the time than two overtakes" —LastJustIn3\\
Full-Model Tuning & NEW students panicked when they heard explosions -- caused by a faulty electrical cable -- and believed it was a terror attack, university says . As many as 108 students from University of Nairobi were admitted to Kenyatta National Hospital . One person died, according to the school .\\
Reference & Students stampeded; some jumped from a fifth story at a dorm; one student died, school officials say . The blasts were caused by faulty electrical cable, and Kenya Power is at the school . The panic came less than two weeks after terrorists attacked Kenya's Garissa University .\\\bottomrule
\end{tabular}}

\caption{Qualitative examples of CNNDM.}
\label{tab:cases cnndm}
\end{table*}

\begin{table*}
\centering
\resizebox{\linewidth}{!}{
\begin{tabular}{lp{17cm}}
\toprule
Source & Brunon Kwiecien, 48, was convicted of planning a terrorist attack, illegal weapons possession and inciting two students to carry out an attack.He suggested he had been manipulated by Polish intelligence agents.Kwiecien was said to be fascinated with Norwegian mass killer Anders Behring Breivik.Right-wing extremist Breivik killed 77 people in a bombing and shooting rampage in Norway in July 2011.Kwiecien, a former professor at Krakow's University of Agriculture, was arrested in 2012.Investigators believe he wanted to target parliament with four tonnes of explosives while then-President Bronislaw Komorowski and former Prime Minister Donald Tusk were due to attend, the trial heard."If Brunon Kwiecien hadn't been stopped, we would be talking amid the ruins of the state today," said judge Aleksandra Almert, according to the AFP agency.While admitting he planned to carry out the attack, he also said he was subject to "provocation" by the intelligence services.Kwiecien is the first Pole to be accused of terrorism, Polish media reported. He has no known links to established extremist groups.     \\ \midrule
PSP  & A Pole has been convicted of planning a terrorist attack in Poland, a court heard. \\
Prefix-Tuning  & A Pole has been convicted of planning to carry out a terrorist attack in Poland. \\
Prompt Tuning &  AA Polish man has been convicted of planning a terrorist attack in the Polish capital, Warsaw, on Thursday. \\
Full-Model Tuning & A Pole has been found guilty of planning a terrorist attack in the Polish state of Krakow.\\
Reference & A Polish university lecturer has been sentenced to 13 years in jail for plotting to ram a car packed with explosives into parliament.\\ \toprule
Source  & Schmidt was sent off by the referee for insulting Hoffenheim's Julian Nagelsmann in Saturday's 3-0 home loss."That was nothing, what sort of a nutcase are you? Just shut your mouth," Schmidt shouted after going 2-0 down.The 49-year-old has been banned for two games and handed a 15,000 euros (£13,373) fine.The German was sanctioned after triggering a suspended sentence from February this year.He had been banned for three games, with a further two in the event of a repeat offence before June 2017, for refusing a referee's order to leave the sidelines during a 1-0 defeat to Borussia Dortmund.Schmidt will be unable to have any contact with the team for half an hour before, during and after Tuesday's German Cup second-round match against Lotte and Saturday's league match against Wolfsburg.Leverkusen's director of sport Rudi Voller has sought a meeting with the head of the disciplinary committee.       \\ \midrule
PSP   &  Leverkusen defender Christian Schmidt has been banned for two games for insulting the referee.\\
Prefix-Tuning & Leverkusen midfielder Matthias Schmidt has been banned for two games after refusing to leave the sidelines during a match against Wolfsburg.\\
Prompt Tuning & ALeverkusen midfielder Christian Schmidt has been banned for two games for insulting the referee in a game against Hoffenheim on Saturday..'\\
Full-Model Tuning & Aeverkusen manager Gerhard Schmidt has been banned for two games for insulting the head of the German national team.\\
Reference &  Bayer Leverkusen head coach Roger Schmidt has been banned and fined for calling an opposing manager "a nutcase" during a Bundesliga game.\\\bottomrule
\end{tabular}}

\caption{Qualitative examples of XSum.}
\label{tab:cases xsum}
\end{table*}

\end{document}